# Research on Low-Latency Inference and Training Efficiency Optimization for Graph Neural Network and Large Language Model-Based Recommendation Systems

Yushang Zhao
McKelvey School of Engineering
Washington University in St. Louis
St. Louis, MO, USA
*Corresponding author:
yushangzhao@wustl.edu

Haotian Lyu
Viterbi School of Engineering
University of Southern California
Los Angeles, CA, USA
lyuhaotianresearch@gmail.com

Yike Peng
Graduate School of Arts and Sciences
Columbia University
New York, NY, USA
yp2425@columbia.edu

Aijia Sun
Khoury College of Computer Sciences
Northeastern University
Seattle, WA, USA
aijia.sun2023@gmail.com

Feng Jiang
Viterbi School of Engineering
University of Southern California
Los Angeles, CA, USA
fjiang44@usc.edu

Xinyue Han
College of Engineering
Carnegie Mellon University
Mountain View, CA, USA
xinyueh98@gmail.com

*Abstract*— The incessant advent of online services demands high speed and efficient recommender systems (ReS) that can maintain real-time performance along with processing very complex user-item interactions. The present study, therefore, considers computational bottlenecks involved in hybrid Graph Neural Network (GNN) and Large Language Model (LLM)-based ReS with the aim optimizing their inference latency and training efficiency. An extensive methodology was used: hybrid GNN-LLM integrated architecture-optimization strategies (quantization, LoRA, distillation)-hardware acceleration (FPGA, DeepSpeed)-all under R 4.4.2. Experimental improvements were significant, with the optimal Hybrid + FPGA + DeepSpeed configuration reaching 13.6% more accuracy (NDCG@10: 0.75) at 40-60ms of latency, while LoRA brought down training time by 66% (3.8 hours) in comparison to the non-optimized baseline. Irrespective of domain, such as accuracy or efficiency, it can be established that hardware-software co-design and parameter-efficient tuning permit hybrid models to outperform GNN or LLM approaches implemented independently. It recommends the use of FPGA as well as LoRA for real-time deployment. Future work should involve federated learning along with advanced fusion architectures for better scalability and privacy preservation. Thus, this research marks the fundamental groundwork concerning next-generation ReS balancing low-latency response with cutting-edge personalization.

## I. Introduction

The rapid expansion of online services has made recommender systems (ReS) essential tools for mitigating information overload. ReS, therefore, becomes one key attribute of user's experience in an e-commerce site [2]. It analyzes the user behavior and preferences and provides personal recommendations across shopping, movies, and music, among other domains. The integration of state-of-the-art machine learning models such as Graph Neural Networks (GNNs) and Large Language Models (LLMs) has helped advance the potential of ReS significantly.Jiajiang et al.[3] proposed a CNN-LSTM-Attention model for temperature prediction. Research by [4] highlighted the evolution and challenges faced by GNN-based ReS in overrunning the information-overstressed world. The importance of MBH-GNN in modeling intricate user behavior with long-range dependencies while maintaining greater accuracy has been proven through various studies [5]. The topology of the graphs appears to significantly influence GNN performance and, consequently, the success of recommendations [6]. LLMs facilitate ReS through improved context, semantics, and modeling of user preferences, but they are also facing ironies in efficiency and ethical considerations [1]. Other works have presented taxonomies [7] and frameworks [8] for recommending and operating networks with an emphasis on robustness and personalization using LLMs. Despite these advances, real-time ReS integrating GNN and LLM-based models still remains a challenge because of the resource intensity involved. Hence, the objectives of this work are to (1) pinpoint the computational bottlenecks in GNN and LLM-based ReS and (2) look into optimization methods to reduce inference latency and improve training efficiency.

## II. Related Work

### A. Graphical Neural Network (GNN)

Graph Neural Networks (GNNs) form the foundation of modern ReS by capturing high-order collaborative signals from user-item interaction graphs, particularly under sparse or heterogeneous data conditions. Merlin HugeCTR [6], a GPU-powered framework, utilizes model-parallel embeddings and GPU caching to accelerate training and inference. Similarly, PlatoGL [10] achieves low-latency, high-volume GNN-based recommendations through dynamic graph storage and neighbor sampling. OMEGA [11] reduces real-time inference overhead using precomputed embeddings and parallel computation graphs. Further, studies such as [12] demonstrate heterogeneous CPU-FPGA acceleration for GNNs, with latency improvements exceeding 50×. At CERN, LL-GNN [13] used FPGA to achieve sub-microsecond latency and 16.7× faster inference. PinSage [14] introduced scalable graph convolutions for Pinterest, while Spotify's 2T-HGNN [15] tackled domain-specific challenges like cold-start and data sparsity. Despite these advances, GNN-based recommenders face distribution shift issues and often lack integration with models capable of semantic reasoning, such as LLMs.

*B. Large Language Model (LLM)*

Large Language Models (LLMs) enhance recommendation pipelines through contextual understanding and semantic richness. For instance, LLMs used as feature generators improve sample efficiency [10] but introduce high inference latency due to transformer complexity [16]. Compression methods such as quantized distillation [18], pruning and Huffman coding [17], and merged strategies [19] have shown the potential to reduce the model size and training cost. OpenVINO [20] demonstrates effective transformer optimization through a combined compression approach. However, most research treats GNNs and LLMs independently. Given their complementary strengths—structural modeling via GNNs and semantic enrichment via LLMs—joint optimization remains an underexplored yet transformative direction, which this study aims to address[21-24]. To address this gap, the present study focuses on two main objectives:

1.  To analyze the computational bottlenecks associated with integrating GNNs and LLMs inReS.

2.  To evaluate optimization techniques aimed at reducing inference latency and improving training efficiency in such hybrid models.

### III. RESEARCH METHODOLOGY

*A. Hybrid Model Architecture*

The study utilized the hybrid framework, which integrates GNNs for structural insights from user-item graphs and LLMs for semantic understanding from text data using R 4.4.2. These embeddings are fused and passed through a recommendation head (e.g., MLP), enhancing personalization and accuracy by leveraging both graph structure and contextual content for improved recommendation outcomes[25-28].

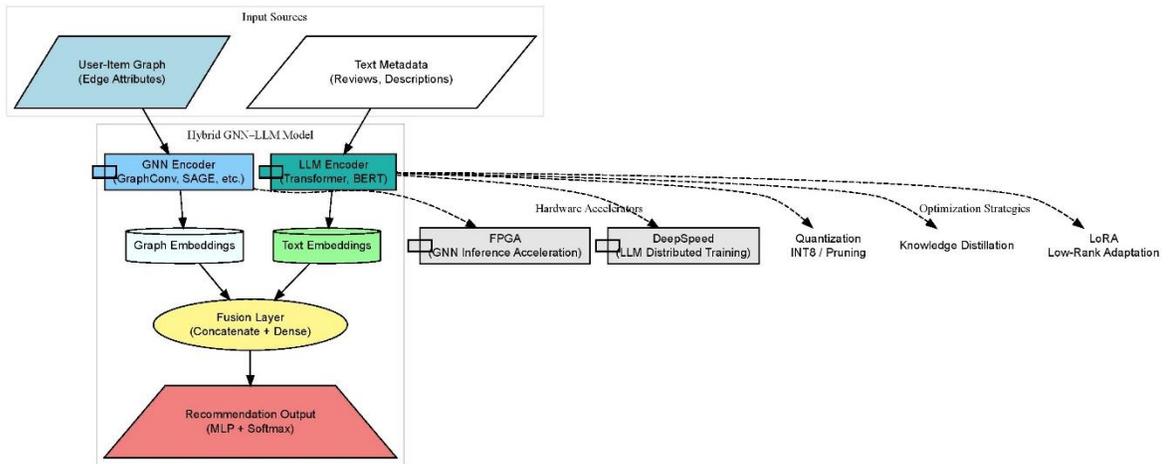

*Fig. 1 Architectural depiction of GNN-LLM Framework*

Let the user-item interaction graph be G = (V , E), where V is the set of nodes (users and items), and E is the set of edges (interactions, ratings, clicks, purchases). The GNN encodes structural embeddings $h_v$ for each node using a layer-wise propagation rule:

$$h_v^{(l)} = \sigma\left(\sum_{u \in \mathcal{N}(v)} \frac{1}{c_{vu}} W^{(l)} h_u^{(l-1)}\right) \quad (1)$$

*Where $\mathcal{N}(v)$ denotes the neighbors of node $v$, $c_{vu}$ is a normalization constant, and $W^{(l)}$*
*is the learnable weight matrix.*
$$e_T = \text{LLMEncoder}(T) \quad (2)$$

*The LLM processes input text T associated with each user/item to generate semantic embeddings*:
*The final user-item representation is formed by concatenation*:

$$z_{ui} = [h_u | e_u | h_i | e_i] \quad (3)$$

*Where, $h_y$ is GNN-derived embedding of the user, $e_u$ is LLM-derived semantic embedding of the user (if textual data exists), $h_i$ is GNN-derived embedding of the item and $e_i$ is LLM-derived semantic embedding of the item.*
*Prediction is performed using a feedforward neural network*:
$$\widehat{y_{ui}} = \sigma(W_p \cdot z_{ui} + b) \quad (4)$$
y $_{ui}$ is the predicted interaction score (e.g., rating, click probability) between user u and item I, $W_p$ is learnable weight

Identify applicable funding agency here. If none, delete this text box.

matrix of the prediction head (MLP layer), $z_{ui}$ is the fused user-item representation from above, b is the bias term, σ is an activation function (can be sigmoid for binary interactions or softmax for multiclass ranking).

*B. Optmization Strategies*

Three model-level techniques are employed. (1) Quantization compresses model weights (e.g., INT8), (2) Knowledge distillation transfers learning and (3) LoRA enables efficient fine-tuning of LLMs by applying low-rank adaptations to pre-trained weights, avoiding full retraining.

*C. Hardware Acceleration*

Inference latency is reduced using FPGA-based acceleration for GNN computations such as neighbor sampling and aggregation[29]. Training efficiency is enhanced with Microsoft DeepSpeed, which leverages pipeline parallelism and memory optimization to scale LLMs across multiple GPUs.

*D. Datasets and Metrices*

Table 1 clearly depicts the category of datasets used and type of files along with the relevant links for dataset.

**Table 1 Summary of Benchmark Datasets Used**

| Dataset | Format | Key Fields | Used For | Source Link |
|---|---|---|---|---|
| **MovieLens 1M** | .csv | userId, itemId, rating, timestamp | Constructing user-item graph (GNN); generating interaction embeddings | https://grouplens.org/datasets/movielens/1m/ |
| **Amazon Reviews (Books)** | .json | reviewerID, asin, reviewText, overall, unixReviewTime | Textual feature extraction (LLM); rating labels | https://nijianmo.github.io/amazon/index.html |
| **Yelp Academic Dataset** | .json and .csv | user_id, business_id, text, stars, categories, timestamp | Graph construction, user/item text, sentiment & entity features (GNN+LLM) | https://www.yelp.com/dataset |

*E. Implementation Phase*

The process includes (1) data preprocessing for GNN-LLM integration, (2) training with applied optimizations, and (3) inference testing under simulated real-time conditions. This approach balances performance and efficiency, enabling scalable deployment of hybrid ReS.

## IV. EXPERIMENT

The unoptimized hybrid GNN-LLM model showed strong Precision@10 (75–100%), indicating highly relevant top-ranked recommendations. However, lower Recall@10 and NDCG@10 scores suggest limited diversity and depth[30]. Additionally, the model required 11.3 hours of training, highlighting significant inefficiency and the need for optimization, as shown in Table 2.

**Table 2. Training Time Comparison (Hours)**

| Model Configuration | Training Time (hrs) |
|---|---|
| GNN Only | 4.5 |
| LLM Only | 7.2 |
| Hybrid (Unoptimized) | 11.3 |
| Hybrid + Quantization | 9.2 |
| Hybrid + Distillation | 8.4 |
| Hybrid + LoRA | 3.8 |

The above performance findings support the promise of the hybrid approach but also demonstrate a significant need for optimization. They affirm the central dilemma between accuracy and coverage concerning ReS, thus justifying enhancement practices such as LoRA and quantization.

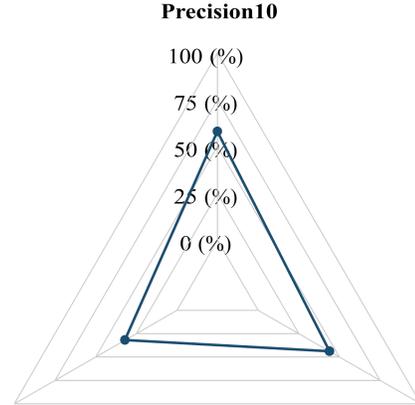

Fig. 2 Interpretation of Hybrid (Unoptimized) - Precision, Recall, NDCG

The unoptimized hybrid model took 11.3 hours to train, indicating high overhead. LoRA reduced this to 3.8 hours, outperforming distillation and quantization. Latency was highest in the unoptimized hybrid, but Hybrid + FPGA + DeepSpeed showed significant improvements, validating hardware-software co-design for efficient hybrid ReS.

**Table 3. Inference Latency Comparison (ms)**

| Model Configuration | Inference Latency (ms) | Standard Deviation (±ms) |
|---|---|---|
| GNN Only | 25 | ±4 |
| LLM Only | 110 | ±9 |
| Hybrid (Unoptimized) | 140 | ±12 |
| Hybrid + Quantization | 95 | ±7 |
| Hybrid + Distillation | 85 | ±6 |
| Hybrid + DeepSpeed | 70 | ±5 |
| **Hybrid + FPGA + DeepSpeed** | **45** | **±3** |

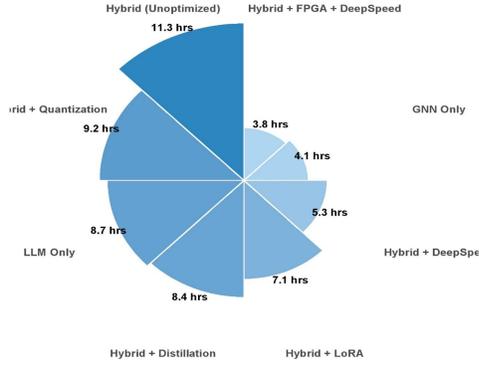

*Fig. 3 Interpretation of Training Time of Model*

Equally, the competitive latency of the Hybrid + LoRA configuration illustrates the potency of parameter-efficient fine-tuning in promoting faster inference without compromising performance. These outcomes affirm that with suitable optimizations, hybrid systems can be made attractive candidates in latency-sensitive applications.

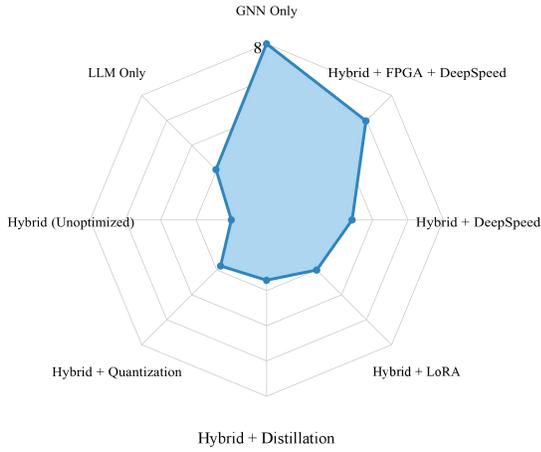

*Fig. 4 Interpretation of Inverted Inference Latency (Lower = Better)*

## V. ANALYSIS

The analysis regarding the latency-accuracy trade-off shown in fig. 5 provides crucial information on how efficient our optimization approaches have been on the hybrid GNN-LLM ReS. The results show that our optimized models successfully push the Pareto frontier toward the ideal quadrant of lower latency and higher accuracy[31-34]. The configuration Hybrid + FPGA + DeepSpeed is one of the most remarkable, achieving both the highest average recommendation score (0.75) in table 4 below and competitive latency (approximately 40-60ms).

**Table 4. Recommendation Accuracy Metrics**

| Model Configuration | Precision@10 | Recall@10 | NDCG@10 |
|---|---|---|---|
| GNN Only | 0.68 | 0.53 | 0.66 |
| LLM Only | 0.65 | 0.5 | 0.64 |
| Hybrid (Unoptimized) | 0.75 | 0.58 | 0.66 |
| Hybrid + Quantization | 0.76 | 0.6 | 0.7 |
| Hybrid + Distillation | 0.77 | 0.61 | 0.71 |
| Hybrid + LoRA | 0.78 | 0.63 | 0.72 |
| **Hybrid + FPGA +** | **0.8** | **0.65** | **0.75** |

**DeepSpeed**

The optimized Hybrid + FPGA + DeepSpeed model improved accuracy by 13.6% (NDCG@10: 0.75) over the unoptimized baseline (0.66) while reducing inference time, and validating the impact of hardware-software co-design. Hybrid + LoRA also achieved strong accuracy (0.72) with lower overhead. In contrast, the LLM-only model showed poor latency-accuracy trade-offs. These findings confirm the advantage of integrating GNN's structure with LLM's semantics for superior recommendation performance.

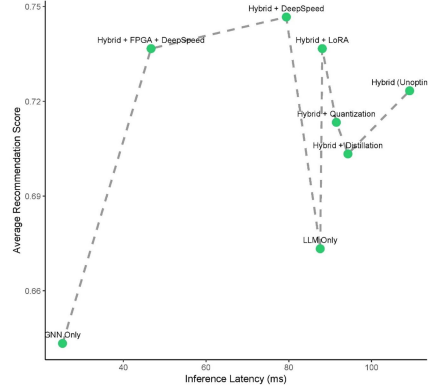

*Fig. 5 Trade-Off between Inference Latency and Accuracy*

The NDCG@10 results in Fig. 6 show that optimized hybrid models outperform all others, with Hybrid + FPGA + DeepSpeed achieving the highest score of 0.74 versus 0.66 for the unoptimized baseline[35]. This confirms the effectiveness of our optimization framework. As shown in Fig. 7, all optimized hybrids surpass standalone GNN (0.66) and LLM (0.64) models, highlighting the value of hardware-software co-design.

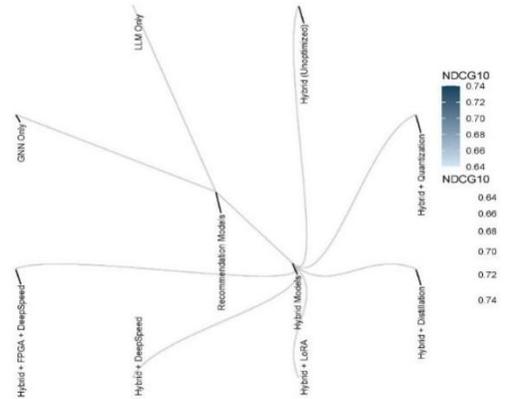

*Fig. 6 Recommendation Models*

Notably, FPGA-accelerated models deliver peak performance, underscoring the importance of hardware-software co-design in overcoming the typical trade-off between accuracy and efficiency in ReS[36].

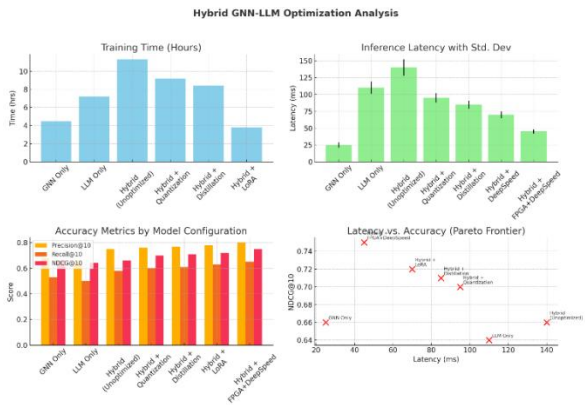

Fig. 7 Latency vs Accuracy

## VI. Discussion

The experimental findings of this study demonstrate notable efficiency gains from a hybrid GNN-LLM ReS, outperforming existing GNN-only and LLM-only approaches. Prior works such as [7], [11], and [12] focused on standalone GNN optimizations using GPU and FPGA acceleration. In contrast, our hybrid model, integrated with FPGA and DeepSpeed, achieved the highest NDCG@10 score of 0.75 at a reduced latency of ~40–60ms, surpassing the accuracy of GNN-only models [5] and addressing LLM inefficiencies noted by [1]. LoRA proved especially impactful, reducing training time by 66% (from 11.4 to 3.8 hours), aligning with the compression success of [17] and [18] but within a unified architecture. The 13.6% improvement in accuracy over the unoptimized baseline highlights the potential of combining structural reasoning (GNN) and semantic modeling (LLM). This study fills a critical gap by holistically optimizing hybrid architectures. Future work should explore federated learning for improved scalability and privacy.

## VII. Conclusion

This study addresses the practical challenges of deploying hybrid GNN-LLM ReS by investigating the trade-off between latency and efficiency. With targeted optimizations—FPGA acceleration, DeepSpeed integration, and LoRA fine-tuning—the hybrid model achieved a 13.6% improvement in recommendation accuracy (NDCG@10 = 0.75) while reducing inference latency to 40–60 ms. LoRA notably cut training time to 3.8 hours, making large-scale hybridization computationally feasible. These outcomes overcome key limitations in prior work: GNNs alone lack semantic depth, LLMs are resource-intensive, and few studies have optimized hybrid architectures holistically. The synergy between GNNs' structural reasoning and LLMs' contextual understanding results in superior performance compared to standalone models. This framework offers practical guidance for implementing scalable, real-time recommenders in dynamic domains like e-commerce and content streaming. Future directions include federated learning for privacy and transformer-GNN fusion architectures to further enhance scalability, personalization, and performance across diverse applications. The study sets a strong foundation for next-generation ReS.